\documentclass[runningheads]{llncs}
\usepackage{amsmath}
\usepackage[T1]{fontenc}
\usepackage{xcolor}
\usepackage{graphicx}
\usepackage{array}
\usepackage{algorithm}
\usepackage{algpseudocode}

\usepackage{tikz,graphics,color,fullpage,float,epsf,caption,subcaption}

\usepackage{multirow}

\usepackage{caption} 
\newcommand*\samethanks[1][\value{footnote}]{\footnotemark[#1]}

\begin{document}
\title{DISBELIEVE: Distance Between Client Models is Very Essential for Effective Local Model Poisoning Attacks}

\title{DISBELIEVE: Distance Between Client Models is Very Essential for Effective Local Model Poisoning Attacks}
%
%

\author{Indu Joshi\inst{1}\thanks{These authors contributed equally to this work.}\and
Priyank Upadhya\inst{1}\samethanks \and
Gaurav Kumar Nayak\inst{2}\and
Peter Schüffler\inst{1}\and
Nassir Navab\inst{1}}

%
\institute{Technical University of Munich, Boltzmannstraße 15, 85748 Garching bei München, Germany \and
Center for Research in Computer Vision, University of Central Florida, Orlando, FL 32816, USA
\email{\{indu.joshi,priyank.upadhya,peter.schueffler,nassir.navab\}@tum.de}\\
\email{\{gauravkumar.nayak\}@ucf.edu}
}

\maketitle              
\begin{abstract}
Federated learning is a promising direction to tackle the privacy issues related to sharing patients' sensitive data. Often, federated systems in the medical image analysis domain assume that the participating local clients are \textit{honest}. Several studies report mechanisms through which a set of malicious clients can be introduced that can poison the federated setup, hampering the performance of the global model. To overcome this, robust aggregation methods have been proposed that defend against those attacks. We observe that most of the state-of-the-art robust aggregation methods are heavily dependent on the distance between the parameters or gradients of malicious clients and benign clients, which makes them prone to local model poisoning attacks when the parameters or gradients of malicious and benign clients are close. Leveraging this, we introduce DISBELIEVE, a local model poisoning attack that creates malicious parameters or gradients such that their distance to benign clients' parameters or gradients is low respectively but at the same time their adverse effect on the global model's performance is high. Experiments on three publicly available medical image datasets demonstrate the efficacy of the proposed DISBELIEVE attack as it significantly lowers the performance of the state-of-the-art 
\textit{robust aggregation} methods for medical image analysis. Furthermore, compared to state-of-the-art local model poisoning attacks, DISBELIEVE attack is also effective on natural images where we observe a severe drop in classification performance of the global model for multi-class classification on benchmark dataset CIFAR-10.

\end{abstract}

\section{Introduction}
The success of deep models for medical image analysis \cite{app13148007} greatly depends on sufficient training data availability. Strict privacy protocols and limited availability of time and resources pose challenges in collecting sizeable medical image datasets \cite{10.1007/978-3-319-41501-7_64}. Although different medical institutions may be willing to collaborate, strict privacy protocols governing patients' information restrict data sharing. Federated learning (FL) offers a promising solution that allows different institutions to share information about these models without revealing personal information about the patients \cite{sheller2020federated,mcmahan2023communicationefficient,dayan2021federated}. Federated Learning is a machine learning paradigm that learns a single shared global model by collaboratively learning from different local models on distributed systems without sharing the data. 

\par A federated learning setup involves multiple clients and a global server \cite{mcmahan2023communicationefficient}. The global server initializes the global model and sends the parameters back to the clients. The clients then train their local models on the data present locally. Once the local models are trained, the parameters are sent to the global model for aggregation. The global model then uses an \textit{aggregation algorithm} to aggregate all the parameter updates and transmits the updated parameters back to the clients, and the cycle repeats until convergence. The federated learning setup allows the clients to preserve the privacy of their data.
 The success of a federated learning system is majorly dependent on the use of an aggregation algorithm. For example, \textit{Federated Averaging} \cite{mcmahan2023communicationefficient} is an aggregation algorithm in which all the parameters accumulated at the global server from different clients are averaged. However, not all clients would act truthfully in real-world scenarios, and there may be some \textit{byzantine} clients. A client is said to be a byzantine client if it acts malicious intentionally due to the presence of an adversary or unintentionally due to faulty equipment or hardware issues \cite{yin2021byzantinerobust}. Studies report that even a single byzantine worker can seriously threaten the FL systems \cite{blanchard2017byzantinetolerant}. 
 
 \par A malicious byzantine worker with an adversary who knows the client's data and model parameters can induce \textit{local poisoning attacks} to degrade the performance of the global model in an FL system.
 A local poisoning attack in an FL system is a process through which the training of the global model is adversely affected due to either data perturbation or perturbation in model parameters (or gradients) at the local client's side. These attacks are termed as \textit{local data poisoning attacks} or \textit{local model poisoning attacks}, respectively. Several studies indicate that state-of-the-art aggregation methods, for instance, using federated averaging in the presence of a byzantine client, will reduce the performance of the global server. Therefore, to defend against attacks by byzantine clients, the global server uses \textit{robust aggregation algorithms} \cite{yin2021byzantinerobust,xie2018generalized}. This research studies the efficacy of state-of-the-art robust aggregation methods for FL systems for medical image analysis and highlights their vulnerability to local model poisoning attacks. We observe that the state-of-the-art robust aggregation methods heavily rely on the distance between malicious and benign client model parameters (or gradients). We argue that some model poisoning attacks can exist when the parameters or gradients of malicious clients are close in Euclidean space to those of benign clients that circumvent the existing state-of-the-art robust aggregation methods. 

 \par \textbf{Research Contribution:} We introduce the DISBELIEVE attack that demonstrates the limitation of state-of-the-art robust aggregation methods for FL on medical images in defending against local model poisoning attacks. The novelty of the proposed attack lies in the fact that it maximizes the objective loss function while ensuring that the Euclidean distance between the malicious parameters and benign parameters is kept marginal. As a result, the attacker can optimally reduce the global model's performance without being detected by the aggregation algorithms. Experiments on three publicly available datasets of different medical image modalities confirm the efficacy of DISBELIEVE attack in significantly reducing the classification performance of the global model (by up to 28\%). We also benchmark two current state-of-the-art local model poisoning attack methods and demonstrate that the proposed DISBELIEVE attack is stronger, leading to higher performance degradation. Lastly, we demonstrate that DISBELIEVE attack also effectively works on natural images, as similar trends are reported on the CIFAR-10 dataset.
 


\section{Related Work}

\subsection{Robust Aggregation Algorithms}
Robust aggregation algorithms are defense methods that prevent malicious clients from significantly affecting parameter updates and global model performance. KRUM \cite{NIPS2017_f4b9ec30} is among the earliest methods for robust aggregation and proposes that for each communication round, only one of the clients is selected as an honest participant, and updates from the other clients are discarded. The client that is chosen as honest is the one whose parameters are closer in Euclidean space to a chosen number of its neighbors. On the other hand, Trimmed Mean \cite{yin2021byzantinerobust} assumes malicious clients to have extreme values of parameters and proposes to avoid malicious clients by selecting parameters around the median. Recently, the Distance-based Outlier Suppression (DOS) \cite{alkhunaizi2022suppressing} algorithm was proposed to defend against byzantine attacks in FL systems for medical image analysis. DOS proposes to detect malicious clients using COPOD, a state-of-the-art outlier detection algorithm \cite{Li_2020}. Subsequently, it assigns less weight to the parameters from those malicious clients. Specifically, it uses Euclidean and cosine distances between parameters from different clients and computes an outlier score for each client. Later, these scores are converted to weights by normalizing them using a softmax function. We note that all these state-of-the-art robust aggregation algorithms assume that malicious clients' parameters (or gradients) are significantly different from benign clients' parameters (or gradients). However, we hypothesize that an attack can be introduced such that parameters (or gradients) of malicious and benign clients are only marginally different, while it can still severely degrade the global model's performance.

\subsection{Attacks in Federated Learning}
There are various kinds of attacks in a federated learning paradigm, such as \textit{inference attacks, reconstruction attacks, poisoning attacks} \cite{chen2022federated,lyu2020threats,9308910}. In inference attacks, the attacker can extract sensitive information about the training data from the learned features or parameters of the model, thus causing privacy issues. Reconstruction attacks, on the other hand, try to generate the training samples using the leaked model parameters \cite{chen2022federated}. GAN's \cite{goodfellow2014generative} have successfully extracted private information about the client's data even when model parameters are unclear due to the use of differential privacy \cite{hitaj2017deep}. Poisoning attacks in a federated learning paradigm can be categorized as \textit{data poisoning attacks} or \textit{model poisoning attacks}. Both these attacks are designed to alter the behavior of the malicious client's model \cite{inbook}. In data poisoning attacks, the attacker tries manipulating the training data by changing the ground truth labels or carefully poisoning the existing data \cite{10.1007/978-3-030-58951-6_24}. In model poisoning attacks, the attacker aims to alter the model parameters or gradients before sending them to the global server \cite{inbook}. 

In this research, we design a model poisoning attack that can bypass state-of-the-art robust aggregation algorithms such as DOS, Trimmed Mean, and KRUM. We evaluate the performance of existing state-of-the-art model poisoning attacks such as LIE attack \cite{baruch2019little} and Min-Max attack \cite{shejwalkar2021manipulating}. We note that the LIE attack forces the malicious parameters (or gradients) to be bounded in a range $(\mu-z\sigma, \mu+z\sigma)$ where $\mu$ and $\sigma$ are the mean and standard deviation along parameters of the malicious clients, and $z$ is a parameter that sets the lower and upper bounds for deviation around the mean \cite{baruch2019little}. On the other hand, Min-Max adds deviation to parameters or gradients and then scales them such that their distance from any other non-malicious parameter is less than the maximum distance between two benign updates. However, instead of relying on standard deviation to approximate the range across which malicious clients' parameters (or gradients) can be manipulated, the proposed attack computes the malicious parameters (or gradients) by maximizing the classification loss (as opposed to minimizing it) to degrade the global model's performance. Additionally, we propose to approximate the range across which the parameters (or gradients) can be perturbed by evaluating the distance between the malicious clients' parameters (or gradients) in Euclidean space.

\begin{figure}[h]
  \centering
  \begin{subcaptiongroup}
    \centering
    \parbox[b]{.4\textwidth}{%
    \centering
    \includegraphics[width=0.25\textwidth]{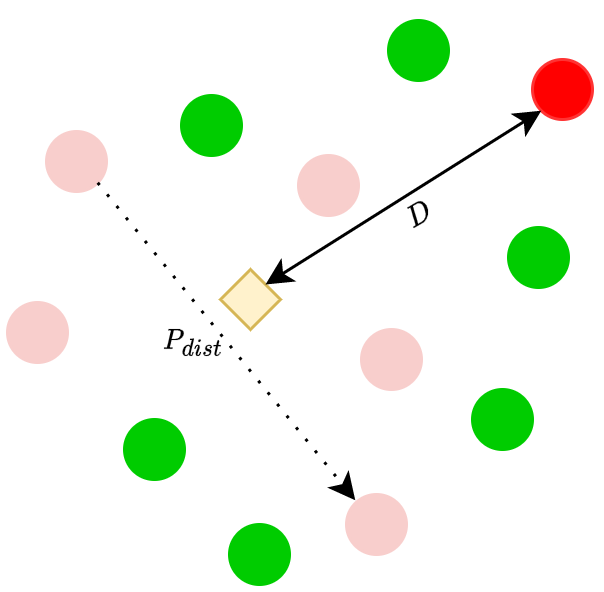}
    \caption{Model Poisoning Attack on Parameters}\label{param_int}}%
    \parbox[b]{.45\textwidth}{%
    \centering
    \includegraphics[width=0.3\textwidth]{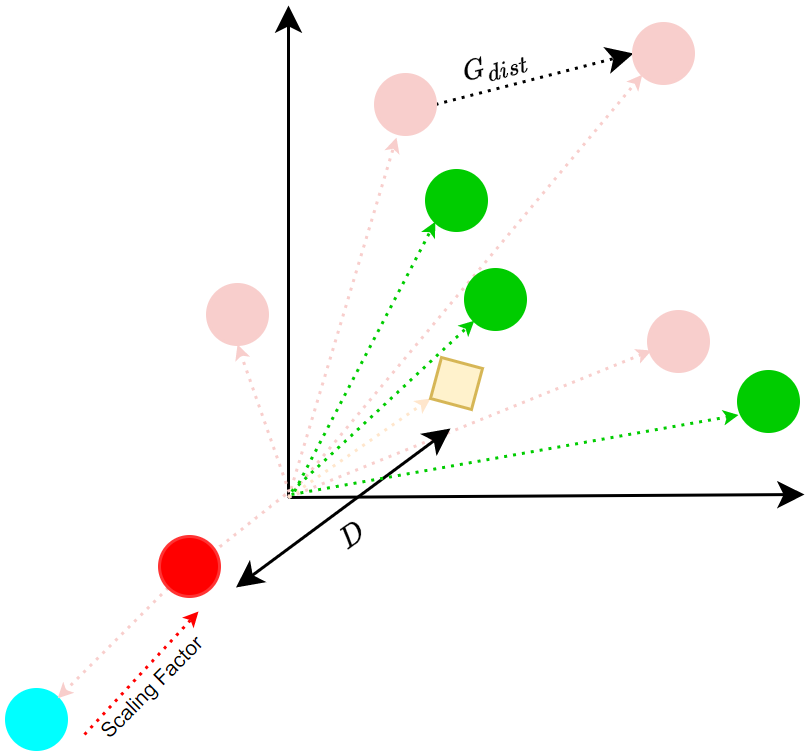}
    \caption{Model Poisoning Attack on Gradient}\label{grad_int}}%
  \end{subcaptiongroup}
  \caption{\textbf{Intuition behind our proposed local model poisoning attack:} (a) Green nodes represent the parameters of benign clients, Pink node represent the parameters of malicious clients, Yellow node represents the mean of malicious clients parameters (i.e average of parameters of Pink nodes), Red node represents the malicious parameters (of model $M$). We ensure that the shift in parameters of model $M$ from mean is less than the threshold $P_{dist}$ where $P_{dist}$ is the maximum distance between any two attacked clients parameters. 
 (b) Green nodes represent gradients of benign clients, Pink nodes represent the malicious clients gradients, Yellow node represents the mean of malicious clients gradients (i.e average gradients of Pink nodes), Blue node represents gradient of trained malicious model $M$, Red node represents gradient of malicious model $M$ after scaling. We ensure that after scaling gradients the distance from mean of gradients is less than threshold  $G_{dist}$ where $G_{dist}$ is the minimum distance between any two attacked clients gradients.}\label{intuition}
\end{figure}

\begin{algorithm}[h]
\small
\caption{DISBELIEVE Attack on Parameters}\label{alg1:model_poisoning_attack_param_agg}
\begin{algorithmic}[1]
    \State Calculate mean of parameters: $$\mu^{param} = \dfrac{1}{f}\sum_{i=1}^{f}W^{mal}_{i}$$
    \State Set the threshold value:$$P_{dist} = Max_{i,k \in f \; i \neq k}||W^{mal}_{i} - W^{mal}_{k}||^{2}_{2}$$
    \State Combine all the training data from malicious clients
    \State Initialize the malicious model $M$ with parameters $\mu^{param}$
    \State Train $M$ with $Loss = -Loss_{class}$ until: $$||W^{mal}_{model} - \mu^{param}||^{2}_{2} \leq P_{dist}$$
    \State Return $W^{mal}_{model}$
\end{algorithmic}
\end{algorithm}

\begin{algorithm}[h]
\small
\caption{DISBELIEVE Attack on Gradients}
\label{alg2:model_poisoning_attack_gradient_agg}
\begin{algorithmic}[1]
    \State Calculate the mean of parameters and gradients: $$\mu^{param} = \dfrac{1}{f}\sum_{i=1}^{f}W^{mal}_{i} \quad\quad \mu^{grad} = \dfrac{1}{f}\sum_{i=1}^{f}Grad^{mal}_{i}$$
    \State Set the threshold value:$$G_{dist} = Min_{i,k \in f \; i \neq k}||Grads^{mal}_i - Grads^{mal}_k||^{2}_{2}$$
    \State Combine all the training data from malicious clients
    \State Initialize the malicious model $M$ with parameters $\mu^{param}$
    \State Train $M$ with $Loss = -Loss_{class}$
    \State $Grads^{mal}_{model} \gets Gradients\;of\;M$
    \State $start \gets 0.001$, $end \gets 1000$
    \While{$|start-end| > 0.01$}
        \State $sf \gets (start+end) / 2$
        \State $Grads^{mal}_{new} = sf * \dfrac{Grads^{mal}_{model}}{||Grads^{mal}_{model}||}$
        \vspace*{1mm}
        \State $diff = ||Grads^{mal}_{new} - \mu^{grad}||^{2}_{2}$
        \State \textbf{if} $diff > G_{dist}$ \textbf{then} $start = sf$ \textbf{else} $end = sf$
    \EndWhile
    \State Return the $Grads^{mal}_{new}$
\end{algorithmic}
\end{algorithm}

\vspace*{-0.75cm}

\section{Proposed Method}
Formally, we assume a total of $n$ federated learning clients out of which $f$ clients ($1 < f < n/2$) have been compromised such that rather than improving global models' accuracy, the compromised clients work towards decreasing the performance of the global model. We further assume that all the attackers corresponding to different malicious clients are working together or that a single attacker controls all the malicious clients. The attacker thus has access to all the malicious client's model parameters and training data. Our goal is to create malicious parameters or gradients that can bypass the robust aggregation algorithms and reduce the performance of the global model. In this direction, this research introduces a model poisoning attack (DISBELIEVE attack) that creates a single malicious model ($M$) with access to parameters, gradients, and training data of all the $f$ clients. $M$ serves as a proxy for $f$ clients and aims towards pushing the output of the global model away from the distribution of the ground truth labels.

To be specific, the malicious model ($M$) is trained to generate malicious parameters or gradients by minimizing the loss $L_{model} = -L_{class}$ as opposed to benign clients where the loss given by $L_{model} = L_{class}$ is minimized. Here $L_{class}$ refers to cross-entropy loss. Once the malicious parameters (or gradients) are computed, $M$ forwards these malicious values to all the $f$ clients, which then transmit these values to the global model. Note that all the $f$ clients receive the same malicious parameters (or gradients) from $M$. Our work leverages the shortcomings of robust federated learning aggregation algorithms such as KRUM \cite{NIPS2017_f4b9ec30} and DOS \cite{alkhunaizi2022suppressing}, which are based on the assumption that malicious parameters or gradients are significantly different from the parameters or gradients of benign clients in euclidean space respectively. Therefore, to reduce the defense capabilities of these aggregation algorithms, it is essential to perturb the parameters (or gradients) so that their Euclidean distance from benign clients' parameters (or gradients) does not become significant. This can be ensured if the Euclidean distance between the malicious parameters (or gradients) and the mean of benign clients' parameters (or gradients) remains bounded. Due to the normal distribution of data, it is safe to assume that the mean of parameters (or gradients) of clients controlled by the attacker is closer to the mean of benign clients parameters (or gradients) respectively in the Euclidean space \cite{baruch2019little}. 

The local model poisoning attack can be introduced on model parameters or gradients \cite{alkhunaizi2022suppressing,baruch2019little}. However, the critical difference between parameters and gradients is that gradients have direction and magnitude, whereas parameters only have magnitude. Hence, we propose different attacks on parameters and gradients. Details on the strategy for attacking parameters or the gradients are provided in Section \ref{par} and Section \ref{grad}, respectively. The attacker initially chooses the clients it wants to attack and accumulates the chosen clients' model parameters, gradients, and training data. Subsequently, the attacker computes the mean of chosen (attacked) clients' model parameters ($\mu^{param}$) and gradients ($\mu^{grad}$) and initializes a new malicious model $M$ with these mean values.
$$\mu^{param} = \dfrac{1}{f}\sum_{i=1}^{f}W^{mal}_{i} \qquad\qquad \mu^{grad} = \dfrac{1}{f}\sum_{i=1}^{f}Grad^{mal}_{i}$$
Here, $W^{mal}_{i}$ and $Grad^{mal}_{i}$ refer to the model parameters or gradients of the $i^{th}$ malicious client respectively.

\subsection{DISBELIEVE attack on Parameters}
\label{par}
The initialized malicious model, $M$, is trained on the accumulated training data for minimizing the loss function $L_{model} = -L_{class}$ until the Euclidean distance between the malicious model's ($M$) parameters and the mean values is less than the maximum distance between any two attacked client's parameters. 
$$||W^{mal}_{model} - \mu^{param}||^{2}_{2} \leq P_{dist} \quad where, \quad P_{dist} = Max_{i,k \in f \; i \neq k}||W^{mal}_{i} - W^{mal}_{k}||^{2}_{2}$$
Here, $W^{mal}_{model}$ refers to the malicious parameters after training of the malicious model $M$, and $P_{dist}$ refers to a threshold. The threshold $P_{dist}$ is critical to ensure a successful attack as it controls how far the malicious parameters can be from the mean of parameters in Euclidean space. Through the proposed attack, we suggest setting this value to the maximum Euclidean distance between any two malicious client parameters. Intuitively this is a reliable value within an upper bound on the malicious parameters by which they can deviate within a fixed bounded Euclidean space around the mean (see Figure \ref{param_int}). The pseudo-code for the attack is given in Algorithm \ref{alg1:model_poisoning_attack_param_agg}.

\subsection{DISBELIEVE attack on Gradients}
\label{grad}

For attacking gradients, as described in Algorithm \ref{alg2:model_poisoning_attack_gradient_agg}, we train the malicious model $M$ with the similar loss function, $Loss = -Loss_{class}$, however, without any thresholding. Once the model $M$ is trained, we accumulate the malicious gradients ($Grads^{mal}_{model}$) and scale them by a scaling factor $sf$ to make sure that their distance from the mean of gradients of malicious clients ($\mu^{grad}$) is smaller than the minimum distance between any two malicious client's gradients ($G_{dist}$) (see Figure \ref{grad_int}). $$G_{dist} = Min_{i,k \in f \; i \neq k}||Grads^{mal}_i - Grads^{mal}_k||^{2}_{2}$$ To find the optimum scaling factor ($sf$), we use a popular search algorithm known as binary search\cite{shejwalkar2021manipulating}. We initialize a start value of 0.001 and an end value of 1000. An optimal $sf$ is computed using the divide and conquer binary search algorithm in between these values, which makes sure that after scaling the unit gradient vector, its distance to the mean of gradients ($\mu^{grad}$) is less than $G_{dist}$ 
$$||sf * \dfrac{Grads^{mal}_{model}}{||Grads^{mal}_{model}||} - \mu^{grad}||^{2}_{2} \leq G_{dist}$$
For calculating gradients, the minimum distance ($G_{dist}$) is preferred over the maximum distance ($P_{dist}$) when attacking parameters. This preference arises because maximizing the objective loss function results in gradients pointing in the opposite direction compared to the direction of benign gradients. By using the minimum distance, we can prevent malicious gradients from becoming outliers.

\section{Experiments}

\subsection{Datasets}
\textbf{CheXpert-Small}: CheXpert \cite{irvin2019chexpert} is a large publicly available dataset containing over 200,000 chest X-ray images for 65,240 patients. However, consistent with the experimental protocol used by state-of-the-art DOS \cite{alkhunaizi2022suppressing}, we use the smaller version of CheXpert, also known as CheXpert-small, that contains 191,456 X-Ray images of the chest. The dataset contains 13 pathological categories. A single observation from the dataset can have multiple pathological labels. Each sample's pathological label is classified as either negative or positive. Consistent with the state-of-the-art aggregation method DOS \cite{alkhunaizi2022suppressing}, we preprocess all the images by rescaling them to 224$\times$224 pixels using the torchxrayvision library.

\textbf{Ham10000}: Ham10000 \cite{Tschandl_2018} or HAM10k is a publicly available benchmark dataset containing dermatoscopic images of common pigmented skin lesions. It is a multi-class dataset with seven diagnostic categories and 
10000 image samples. As suggested in \cite{alkhunaizi2022suppressing}, we use this dataset to evaluate the model performance in non-iid settings where each image is resized to 128$\times$128.

\textbf{Breakhis}: The breakhis dataset \cite{7312934} is a public breast cancer histopathological database that contains microscopic images of breast cancer tissues. The dataset contains 9109 images from 82 different patients. The images are available in magnifying scales such as 40X, 100X, 200X, and 400X. Each image is a 700$\times$460 pixels sized image, and we rescale each image to 32$\times$32 for our classification task. We use this dataset for binary classification of 400X magnified microscopic images where we classify cancer present in images as either benign or malignant.

\textbf{CIFAR-10}: The Cifar-10 \cite{krizhevsky2009learning} is a popular computer vision dataset that contains 60000 natural images of size 32$\times$32. The dataset contains ten classes, and each class has 6000 images. 50000 images are reserved for training, and 10000 images are used for testing.

\subsection{Experimental Setup and Implementation Details}
 The experimental setup used in this research is consistent with the experimental protocols suggested in \cite{alkhunaizi2022suppressing}. Subsequently, we use Chexpert-Small \cite{irvin2019chexpert} and Ham10k datasets \cite{Tschandl_2018} for parameter-based attacks. Likewise, the CheXpert-small dataset is used to train the Resnet-18 \cite{he2015deep} model with a batch size of 16 for 40 communication rounds, and the number of local epochs is set to 1, whereas the Ham10k dataset is trained on a custom model with two convolutional layers and three fully connected layers with a batch size of 890 for 120 communication rounds and the number of local epochs were set to 3. For both datasets, the number of clients is fixed at 10, the number of attackers is fixed at 4, and the learning rate is set to 0.01.

\par For preserving the privacy of clients and their data, federated learning setups usually share gradients instead of model parameters. Hence, we also evaluate our attack for gradient aggregation on the Breakhis \cite{7312934}. Furthermore, to assess the generalization ability of the proposed DISBELIEVE attack on natural images, we evaluate the proposed DISBELIEVE attack on the CIFAR-10 dataset with a gradient aggregation strategy at the global server. For experiments on Breakhis dataset, VGG-11 \cite{simonyan2015deep} model is trained for binary classification. Training occurs for 200 communication rounds with a batch size of 128 and a learning rate 0.0001. For the CIFAR-10 dataset,  we use the VGG-11 \cite{simonyan2015deep} model with ten output classes for 500 communication rounds with a batch size of 1000 and a learning rate of 0.001. Adam optimizer was used for both datasets. The total number of clients and attackers for both datasets is fixed at 10 and 3, respectively.

\begin{figure}[hbt!]
\centering
    \includegraphics[width=0.7\textwidth]{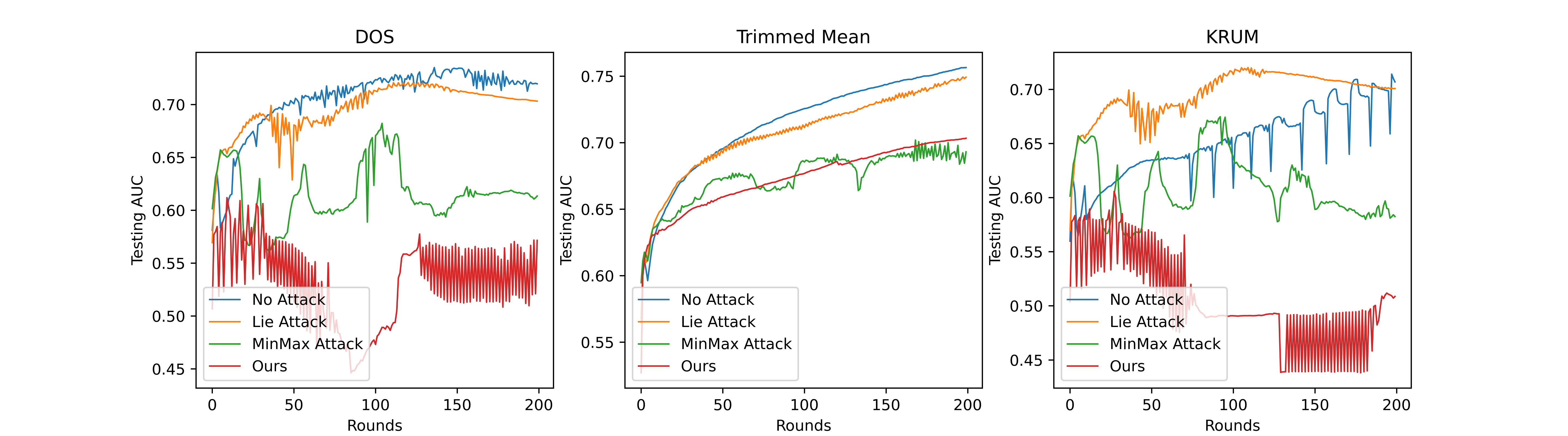}
    \includegraphics[width=0.7\textwidth]{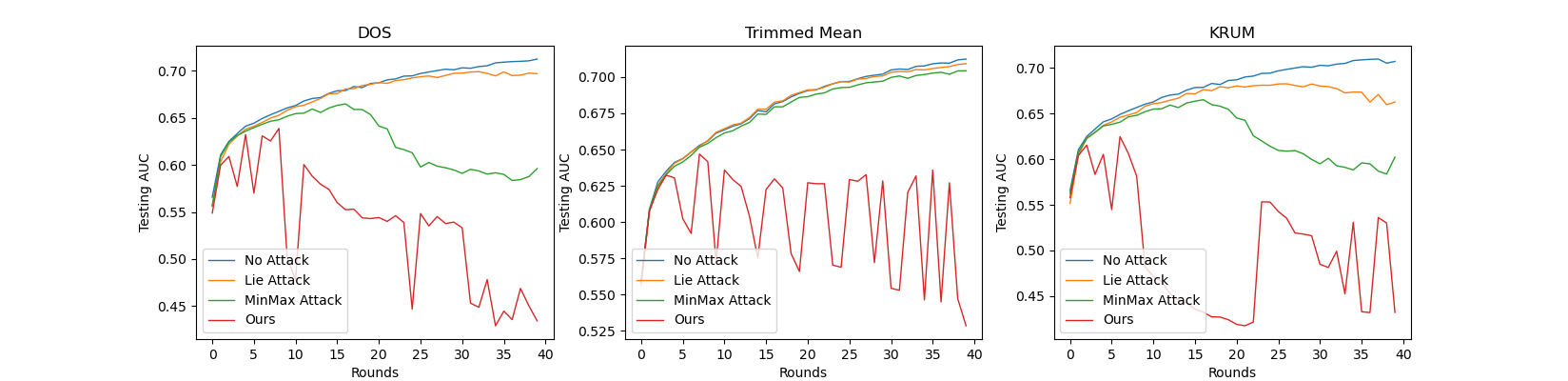}
    \caption{Performance of different attacks on Ham10k (top-row) and CheXpert (bottom-row) datasets under different parameter aggregation methods. Left to right (in order): AUC scores when attacks are made on DOS, Trimmed Mean and Krum.}\label{param_results}
\end{figure}
\begin{figure}[hbt!]
\vspace*{-0.45cm}
\centering
    \includegraphics[width=0.7\textwidth]{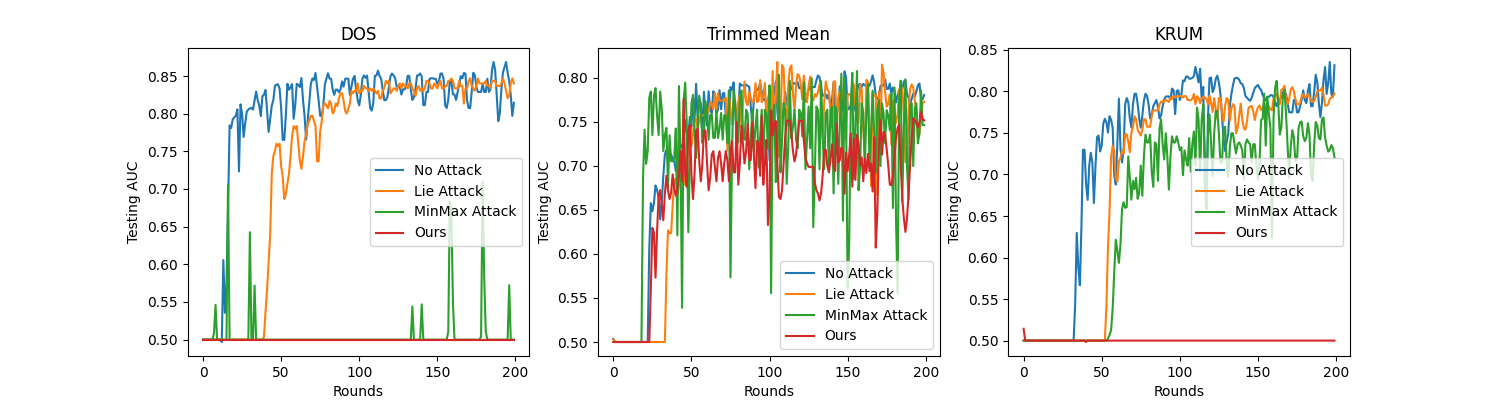}
    \includegraphics[width=0.7\textwidth]{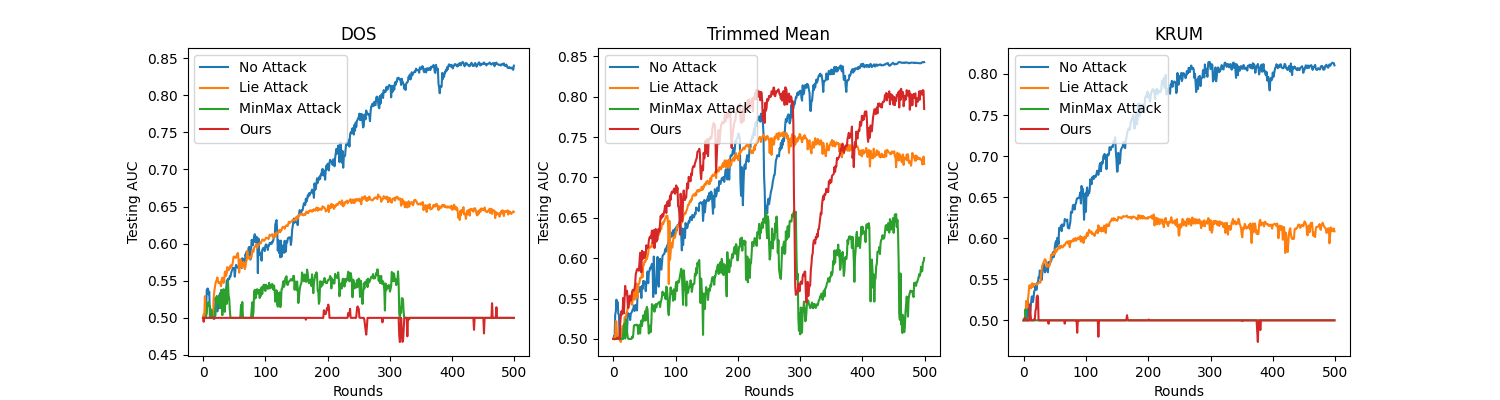}
    \caption{Performance of different attacks on Breakhis (top-row) and CIFAR-10 (bottom-row) datasets under different gradient aggregation methods.  Left to right (in order): AUC scores when attacks are made on DOS, Trimmed Mean and Krum.}\label{grad_results}
\end{figure}

\begin{table}[hbt!]
\caption{Area Under the Receiver Operating Characteristic Curve (AUC) scores with different types of poisoning attack on model parameters}
\centering
\begin{tabular}{>{\centering\arraybackslash}p{3cm}>{\centering\arraybackslash}p{3cm}>{\centering\arraybackslash}p{3cm}>{\centering\arraybackslash}p{3cm}>{\centering\arraybackslash}p{3cm}} 
\hline
Dataset & Attack & DOS & Trimmed Mean & KRUM \\
\hline
\multirow{3}{4em}{Ham10k} & No Attack & 0.72 & 0.75 & 0.70 \\ 
& LIE Attack & 0.70 & 0.74 & 0.70 \\ 
& Min-Max Attack & 0.61 & 0.68 & 0.58 \\
& Ours & 0.52 & 0.70 & 0.51 \\
\hline
\multirow{3}{4em}{CheXpert} & No Attack & 0.71 & 0.71 & 0.70 \\ 
& LIE Attack & 0.69 & 0.71 & 0.65 \\ 
& Min-Max Attack & 0.59 & 0.70 & 0.59 \\
& Ours & 0.44 & 0.52 & 0.43 \\
\hline
\end{tabular}
\label{table_1}
\end{table}

\begin{table}[hbt!]
\vspace*{-0.5cm}
\caption{Area Under the Receiver Operating Characteristic Curve (AUC) scores with different types of poisoning attack on model gradients}
\centering
\begin{tabular}{>{\centering\arraybackslash}p{3cm}>{\centering\arraybackslash}p{3cm}>{\centering\arraybackslash}p{3cm}>{\centering\arraybackslash}p{3cm}>{\centering\arraybackslash}p{3cm}} 
\hline
Dataset & Attack & DOS & Trimmed Mean & KRUM \\
\hline
\multirow{3}{4em}{Breakhis} & No Attack & 0.81 & 0.78 & 0.83 \\ 
& LIE Attack & 0.84 & 0.77 & 0.79 \\ 
& Min-Max Attack & 0.50 & 0.74 & 0.72 \\
& Ours & 0.50 & 0.75 & 0.50 \\
\hline
\multirow{3}{4em}{CIFAR-10} & No Attack & 0.83 & 0.84 & 0.81 \\ 
& LIE Attack & 0.64 & 0.71 & 0.60 \\ 
& Min-Max Attack & 0.50 & 0.60 & 0.50 \\
& Ours & 0.50 & 0.78 & 0.50 \\
\hline
\end{tabular}
\label{table_2}
\end{table}

\section{Results and Discussions}
\subsection{Baselines}
The DISBELIEVE attack is evaluated against three state-of-the-art defense methods: DOS \cite{alkhunaizi2022suppressing}, Trimmed Mean \cite{yin2021byzantinerobust}, and KRUM \cite{NIPS2017_f4b9ec30}. Comparisons are also made with prominent attacks, including LIE \cite{baruch2019little} and Min-Max \cite{shejwalkar2021manipulating}, under different defense methods. Under any defense, AUC scores are highest in the absence of attacks. The LIE attack slightly reduces AUC scores while remaining relatively weaker due to parameter bounding. Conversely, introducing noise and scaling parameters makes the Min-Max attack more potent, consistently reducing AUC scores more significantly across various aggregation methods.

\subsection{Vulnerability of State-of-the-art Defense Methods}
The proposed DISBELIEVE attack reveals the vulnerability of the current state-of-the-art robust aggregation algorithms (Trimmed Mean \cite{yin2021byzantinerobust}, KRUM \cite{NIPS2017_f4b9ec30}, and DOS \cite{alkhunaizi2022suppressing}) over local model poisoning attacks. We empirically validate that our proposed local model poisoning attack (DISBELIEVE attack) can successfully circumvent all three state-of-the-art robust aggregation algorithms (refer Figure \ref{param_results}, Figure \ref{grad_results}). For both parameters and gradient aggregation, DISBELIEVE attack consistently reduces the global model's area under the curve (AUC) scores on all three benchmark medical image datasets. Furthermore, to assess the effectiveness of the proposed DISBELIEVE attack on natural images apart from the specialized medical images, we additionally conduct DISBELIEVE attack on a popular computer vision dataset, CIFAR-10. For natural images, we also find (refer Figure \ref{grad_results}) that the DISBELIEVE attack reduces the global model's AUC score for different state-of-the-art aggregation algorithms DOS, Trimmed Mean, and KRUM. Tables \ref{table_1} and \ref{table_2} show that when subjected to DISBELIEVE attack, the AUC scores fall drastically for all datasets compared to the AUC scores in case of no attack. Therefore, these results demonstrate the vulnerability of state-of-the-art robust aggregation methods to the proposed local model poisoning attack.


\subsection{Superiority of DISBELIEVE attack over State-of-the-art Local Model Poisoning Attacks}
The state-of-the-art robust aggregation algorithm for medical images DOS is only evaluated against additive Gaussian noise, scaled parameter attacks, and label flipping attacks. We additionally benchmark the performance of two state-of-the-art model poisoning attacks, namely Min-Max \cite{shejwalkar2021manipulating} and LIE \cite{baruch2019little} on all the three medical image datasets (refer Figure \ref{param_results} and Figure \ref{grad_results}). Results establish the superiority of the proposed DISBELIEVE attack over state-of-the-art model poisoning attacks on different medical image datasets. While using DOS and KRUM aggregation, the DISBELIEVE attack reduces the global model's AUC score by a more significant margin than both Min-Max and LIE for all the datasets. In the case of trimmed mean, the results of DISBELIEVE attack are comparable on Ham10k (parameter aggregation) and Breakhis (gradient aggregation) datasets with the Min-Max attack and better on CheXpert (parameter aggregation) dataset when compared to the Min-Max and LIE attacks. To compare the effectiveness of DISBELIEVE attack with state-of-the-art model poisoning attacks on the natural image dataset (CIFAR-10), we observe that DISBELIEVE attack performs better than LIE and Min-Max on DOS and KRUM defenses. Tables \ref{table_1} and \ref{table_2} compare state-of-the-art model poisoning attacks and the proposed DISBELIEVE attack under different state-of-the-art robust aggregation algorithms for parameter and gradient aggregation, respectively.

\section{Conclusion and Future Work}
This research highlights the vulnerability of state-of-the-art robust aggregation methods for federated learning on medical images. Results obtained on three public medical datasets reveal that distance-based defenses fail once the attack is designed to ensure that the distance between malicious clients and honest clients' parameters or gradients is bounded by the maximum or minimum distance between parameters or gradients of any two attacked clients, respectively. Moreover, we also demonstrate that the proposed DISBELIEVE attack proves its efficacy on natural images besides domain-specific medical images. In the future, we plan to design a robust aggregation algorithm for federated learning in medical images that can withstand the proposed local model poisoning attack.

\vspace{4mm}
\noindent
\textbf{Acknowledgment.} This work was done as a part of the IMI BigPicture project (IMI945358).

\bibliographystyle{splncs04}
\bibliography{mybibliography}

\begin{thebibliography}{10}
\providecommand{\url}[1]{\texttt{#1}}
\providecommand{\urlprefix}{URL }
\providecommand{\doi}[1]{https://doi.org/#1}

\bibitem{alkhunaizi2022suppressing}
Alkhunaizi, N., Kamzolov, D., Tak{\'a}{\v{c}}, M., Nandakumar, K.: Suppressing
  poisoning attacks on federated learning for medical imaging. In: Medical
  Image Computing and Computer Assisted Intervention--MICCAI 2022: 25th
  International Conference, Singapore, September 18--22, 2022, Proceedings,
  Part VIII. pp. 673--683. Springer (2022)

\bibitem{baruch2019little}
Baruch, M., Baruch, G., Goldberg, Y.: A little is enough: Circumventing
  defenses for distributed learning (2019)

\bibitem{NIPS2017_f4b9ec30}
Blanchard, P., El~Mhamdi, E.M., Guerraoui, R., Stainer, J.: Machine learning
  with adversaries: Byzantine tolerant gradient descent. In: Guyon, I.,
  Luxburg, U.V., Bengio, S., Wallach, H., Fergus, R., Vishwanathan, S.,
  Garnett, R. (eds.) Advances in Neural Information Processing Systems.
  vol.~30. Curran Associates, Inc. (2017)

\bibitem{blanchard2017byzantinetolerant}
Blanchard, P., Mhamdi, E.M.E., Guerraoui, R., Stainer, J.: Byzantine-tolerant
  machine learning (2017)

\bibitem{chen2022federated}
Chen, Y., Gui, Y., Lin, H., Gan, W., Wu, Y.: Federated learning attacks and
  defenses: A survey (2022)

\bibitem{dayan2021federated}
Dayan, I., Roth, H.R., Zhong, A., Harouni, A., Gentili, A., Abidin, A.Z., Liu,
  A., Costa, A.B., Wood, B.J., Tsai, C.S., et~al.: Federated learning for
  predicting clinical outcomes in patients with covid-19. Nature medicine
  \textbf{27}(10),  1735--1743 (2021)

\bibitem{goodfellow2014generative}
Goodfellow, I.J., Pouget-Abadie, J., Mirza, M., Xu, B., Warde-Farley, D.,
  Ozair, S., Courville, A., Bengio, Y.: Generative adversarial networks (2014)

\bibitem{he2015deep}
He, K., Zhang, X., Ren, S., Sun, J.: Deep residual learning for image
  recognition (2015)

\bibitem{hitaj2017deep}
Hitaj, B., Ateniese, G., Perez-Cruz, F.: Deep models under the gan: Information
  leakage from collaborative deep learning (2017)

\bibitem{irvin2019chexpert}
Irvin, J., Rajpurkar, P., Ko, M., Yu, Y., Ciurea-Ilcus, S., Chute, C.,
  Marklund, H., Haghgoo, B., Ball, R., Shpanskaya, K., Seekins, J., Mong, D.A.,
  Halabi, S.S., Sandberg, J.K., Jones, R., Larson, D.B., Langlotz, C.P., Patel,
  B.N., Lungren, M.P., Ng, A.Y.: Chexpert: A large chest radiograph dataset
  with uncertainty labels and expert comparison (2019)

\bibitem{9308910}
Jere, M.S., Farnan, T., Koushanfar, F.: A taxonomy of attacks on federated
  learning. IEEE Security \& Privacy  \textbf{19}(2),  20--28 (2021).
  \doi{10.1109/MSEC.2020.3039941}

\bibitem{10.1007/978-3-319-41501-7_64}
Joshi, I., Kumar, S., Figueiredo, I.N.: Bag of visual words approach for
  bleeding detection in wireless capsule endoscopy images. In: Campilho, A.,
  Karray, F. (eds.) Image Analysis and Recognition. pp. 575--582. Springer
  International Publishing, Cham (2016)

\bibitem{app13148007}
Joshi, I., Mondal, A.K., Navab, N.: Chromosome cluster type identification
  using a swin transformer. Applied Sciences  \textbf{13}(14) (2023).
  \doi{10.3390/app13148007}, \url{https://www.mdpi.com/2076-3417/13/14/8007}

\bibitem{krizhevsky2009learning}
Krizhevsky, A., et~al.: Learning multiple layers of features from tiny images
  (2009)

\bibitem{Li_2020}
Li, Z., Zhao, Y., Botta, N., Ionescu, C., Hu, X.: {COPOD}: Copula-based outlier
  detection. In: 2020 {IEEE} International Conference on Data Mining ({ICDM}).
  {IEEE} (nov 2020). \doi{10.1109/icdm50108.2020.00135},
  \url{https://doi.org/10.1109%2Ficdm50108.2020.00135}

\bibitem{lyu2020threats}
Lyu, L., Yu, H., Yang, Q.: Threats to federated learning: A survey (2020)

\bibitem{inbook}
Lyu, L., Yu, H., Zhao, J., Yang, Q.: Threats to Federated Learning, pp. 3--16
  (11 2020)

\bibitem{mcmahan2023communicationefficient}
McMahan, H.B., Moore, E., Ramage, D., Hampson, S., y~Arcas, B.A.:
  Communication-efficient learning of deep networks from decentralized data
  (2023)

\bibitem{shejwalkar2021manipulating}
Shejwalkar, V., Houmansadr, A.: Manipulating the byzantine: Optimizing model
  poisoning attacks and defenses for federated learning. In: NDSS (2021)

\bibitem{sheller2020federated}
Sheller, M.J., Edwards, B., Reina, G.A., Martin, J., Pati, S., Kotrotsou, A.,
  Milchenko, M., Xu, W., Marcus, D., Colen, R.R., et~al.: Federated learning in
  medicine: facilitating multi-institutional collaborations without sharing
  patient data. Scientific reports  \textbf{10}(1),  1--12 (2020)

\bibitem{simonyan2015deep}
Simonyan, K., Zisserman, A.: Very deep convolutional networks for large-scale
  image recognition (2015)

\bibitem{7312934}
Spanhol, F.A., Oliveira, L.S., Petitjean, C., Heutte, L.: A dataset for breast
  cancer histopathological image classification. IEEE Transactions on
  Biomedical Engineering  \textbf{63}(7),  1455--1462 (2016).
  \doi{10.1109/TBME.2015.2496264}

\bibitem{10.1007/978-3-030-58951-6_24}
Tolpegin, V., Truex, S., Gursoy, M.E., Liu, L.: Data poisoning attacks against
  federated learning systems. In: Chen, L., Li, N., Liang, K., Schneider, S.
  (eds.) Computer Security -- ESORICS 2020. pp. 480--501. Springer
  International Publishing, Cham (2020)

\bibitem{Tschandl_2018}
Tschandl, P., Rosendahl, C., Kittler, H.: The {HAM}10000 dataset, a large
  collection of multi-source dermatoscopic images of common pigmented skin
  lesions. Scientific Data  \textbf{5}(1) (aug 2018).
  \doi{10.1038/sdata.2018.161}, \url{https://doi.org/10.1038%2Fsdata.2018.161}

\bibitem{xie2018generalized}
Xie, C., Koyejo, O., Gupta, I.: Generalized byzantine-tolerant sgd (2018)

\bibitem{yin2021byzantinerobust}
Yin, D., Chen, Y., Ramchandran, K., Bartlett, P.: Byzantine-robust distributed
  learning: Towards optimal statistical rates (2021)

\end{thebibliography}

\end{document}